\documentclass[10pt,twocolumn]{ICCAS}

\usepackage{diagbox}
\usepackage{amsmath}
\usepackage{array}
\usepackage{graphicx}
\usepackage{cite}
\usepackage{amssymb}

\title{Sparse Calibration-Based Personalization of Kernel-Based Gait Phase and Speed Estimation Using Wearable IMUs}

\author{Myeongju Cha$^{1}$ and Pilwon Hur$^{1*}$}

\affils{${}^{1}$Department of Mechanical and Robotics Engineering, GIST, Gwangju, Korea\\
(*Corresponding author: pilwonhur@gist.ac.kr)}

\abstract{
This paper presents sparse calibration-based personalization for kernel-based gait phase and walking speed co-estimation from wearable inertial measurement units (IMUs). A 28-speed reference library was constructed from bilateral thigh and shank trajectories in a 20-subject locomotion dataset. Rather than directly applying population-average kernels, the method combines a user-specific baseline estimated from three calibration speeds with a principal-component (PC) model of baseline-centered kinematic deviations. Offline validation showed that personalized kernels reduced phase error relative to the population kernel. In a single-participant online pilot evaluation, the personalized kernels produced numerically lower mean phase and speed errors and ran in real time on embedded hardware. However, the speed effect was not significant, and pairwise phase differences did not remain significant after multiple-comparison correction. The pilot evaluation establishes wearable implementation feasibility while indicating reference-to-online dataset mismatch and the need for larger-cohort validation.
}


\begin{document}

\maketitle
\thispagestyle{empty}
\pagestyle{empty}

\section{Introduction}

Continuous gait phase and walking speed estimates support wearable robots, powered prostheses, and rehabilitation monitoring~\cite{weigand2022continuous}. Phase describes within-stride timing, whereas speed represents the global locomotor condition. Portable inertial measurement units (IMUs) enable estimation outside motion-capture laboratories~\cite{garcia2023inertial}, but their trajectories may differ from optical measurements because of sensor placement, segment definition, drift compensation, and calibration~\cite{garcia2023inertial,santos2022multi}. This mismatch matters when models learned from motion-capture-derived trajectories are applied to IMU-derived angles.

Continuous phase variables parameterize gait-cycle progression and provide joint-kinematic or impedance references for wearable-robot control. Robust phase estimation under changing locomotion conditions has therefore been studied using mechanical phase variables and phase portraits under perturbations, non-rhythmic tasks, and varying speeds and inclines~\cite{villarreal2017robust,rezazadeh2019phase,best2021phase}. Related approaches address phase-shifting compensation, slopes, stair descent, and speed-dependent walking~\cite{hong2021phase,kumar2022control,cha2025stair,hong2022torso}. Among these sources of variability, this study focuses on inter-user and speed-dependent differences during steady-state level-ground walking.

To estimate gait phase and locomotion condition efficiently, Posh et al.~\cite{posh2025simultaneous} proposed a sliding-kernel framework that selects the best-matching locomotion-mode kernel and its circular alignment. We adapt this principle to phase and speed co-estimation by treating each walking speed as a separate candidate kernel: the selected kernel determines speed, while its alignment determines phase. This approach is interpretable, computationally efficient, and suitable for embedded implementation.

However, matching accuracy depends directly on how well the candidate trajectories represent the current user and sensing modality. Population-average kernels may not capture inter-subject kinematic differences, and motion-capture-derived kernels may not match wearable-IMU measurements. Conversely, constructing a complete user-specific library would require calibration at every candidate speed, which is impractical. To address this limitation, we construct a 28-speed (0.50--1.85 m/s), four-channel reference model from 20 subjects and regenerate user-specific kernels from only three calibration speeds using baseline-centered principal component analysis (PCA) and quadratic PC1-speed regression. Thus, our contribution is not the recursive matching algorithm itself, but sparse personalization of its speed-indexed kernel library.

\section{Kernel-Based Phase and Speed Estimation}

\subsection{Multi-Speed Reference Kernel}

The sliding-kernel framework of Posh et al.~\cite{posh2025simultaneous} jointly estimates locomotion mode and gait phase by selecting the best-matching locomotion-mode kernel and its circular alignment. We adapted this framework to level-ground speed estimation by treating each speed in $\mathcal{V}=\{0.50+0.05r\mid r=0,\ldots,27\}$ m/s as a separate kernel candidate.

Let $x_{s,v,c}(\phi)$ denote the subject-level mean stride trajectory for subject $s$, walking speed $v\in\mathcal{V}$, channel $c$, and normalized gait phase $\phi\in(0,1]$. Four sagittal-plane channels were used: the bilateral thigh and shank angles. After valid strides were resampled to 101 phase samples and averaged within subject and speed, the population kernel was computed as
\begin{equation}
K^{pop}_{v,c}(\phi)
=
\frac{1}{|S(v)|}
\sum_{s\in S(v)}x_{s,v,c}(\phi),
\end{equation}
where $S(v)$ is the set of subjects with valid strides at speed $v$. For matching at 100 Hz, $K^{pop}_{v,c}(\phi)$ was interpolated to the speed-specific kernel length $L_v$, defined as the mean number of samples per stride at speed $v$. The resulting four-channel sample at index $j$ was
\begin{equation}
\mathbf{k}^{pop}_{v,j}
=
\left[
K^{pop}_{v,1}(\phi_{v,j}),\ldots,
K^{pop}_{v,4}(\phi_{v,j})
\right]^{T},
\quad
\phi_{v,j}=\frac{j}{L_v},
\end{equation}
where $j\in\mathcal{J}_v=\{1,\ldots,L_v\}$.

\subsection{Online Matching}

Let $\mathbf{k}_{v,j}$ denote the $j$-th four-channel sample of the population or personalized kernel library being evaluated. For circular indexing, define
\begin{equation}
\langle q\rangle_{L_v}
=
1+\operatorname{mod}(q-1,L_v),
\end{equation}
which maps any integer index to $\mathcal{J}_v$.
At sample $i$, the measured four-channel angle vector $\mathbf{d}_i$ was compared with every candidate speed and circular alignment over the most recent $L_v$ samples:
\begin{equation}
e^{v}_{i,j}
=
\sum_{\ell=0}^{L_v-1}
\left\|
\mathbf{d}_{i-\ell}
-
\mathbf{k}_{v,\langle j-\ell\rangle_{L_v}}
\right\|_2^2.
\end{equation}
The speed and phase estimates were obtained jointly as
\begin{equation}
(\hat{v}_i,\hat{j}_i)
=
\arg\min_{\substack{v\in\mathcal{V}\\j\in\mathcal{J}_v}}
e^{v}_{i,j},
\qquad
\hat{\phi}_i
=
\frac{\hat{j}_i}{L_{\hat{v}_i}}.
\end{equation}
Minimizing the error gives the estimates of speed and normalized phase. Each update was independent of the preceding estimate: no output smoothing, speed-transition penalty, or phase-continuity constraint was used. The implementation also had no matching-cost threshold, confidence measure, or out-of-distribution rejection and therefore always returned an estimate. Temporal context came only from the $L_v$-sample window. For efficiency, the recursive scheme of Posh et al.~\cite{posh2025simultaneous} reused the preceding error array instead of recomputing every complete sliding-window error.

\section{Sparse Personalized Kernel Generation}

\subsection{Baseline-Centered PCA}

The PCA model was constructed from the open-source Camargo locomotion dataset~\cite{camargo2021comprehensive}. Instead of applying PCA to absolute trajectories, each subject's baseline averaged all available speeds:
\begin{equation}
b_s(\phi,c)
=
\frac{1}{|V(s)|}
\sum_{v\in V(s)}
x_{s,v,c}(\phi),
\end{equation}
where $V(s)$ is the set of valid walking speeds for subject $s$. The baseline-centered deviation was defined as
\begin{equation}
\Delta x_{s,v,c}(\phi)
=
x_{s,v,c}(\phi)-b_s(\phi,c).
\end{equation}

For each subject--speed pair, the four-channel deviation trajectories were concatenated over 101 phase samples into a 404-dimensional vector $\Delta\mathbf{x}_{s,v}$. PCA was applied to these vectors collected from all valid subject--speed pairs. Using PC1, each deviation vector was approximated as
\begin{equation}
\Delta\mathbf{x}_{s,v}
\approx
a^{(1)}_{s,v}\mathbf{p}_1,
\end{equation}
where $\mathbf{p}_1$ is the PC1 loading vector and $a^{(1)}_{s,v}$ is the corresponding score. In the full 20-subject model, PC1 explained 84.07\% of the baseline-centered variance and yielded a reconstruction RMSE of 1.371$^\circ$. PC1 was therefore retained as a compact representation, and the association between its scores and walking speed was subsequently quantified through regression.

\subsection{PC1-Speed Regression}

Walking speed was related to PC1 score by polynomial regression. Leave-one-subject-out (LOSO) cross-validation fitted each candidate on 19 subjects and evaluated the held-out subject. The quadratic order had the lowest mean validation RMSE among linear, quadratic, and cubic models. It was refitted on all 20 subjects for online validation:
\begin{equation}
\hat{a}(v)=54.6209v^2-286.0422v+251.7786.
\label{eq:pc1_speed_regression}
\end{equation}

\subsection{Calibration Baseline for a New User}

For offline analysis, one three-speed combination was fixed before performance evaluation using baseline-reconstruction error averaged across all 20 reference subjects. For each subject and candidate combination, an equal-weight calibration baseline was compared with that subject's mean over all 28 speeds. Candidates were restricted to a maximum speed of 1.35 m/s to avoid impractical high-speed calibration. The combination 0.90, 1.15, and 1.35 m/s minimized the mean reconstruction RMSE. This selection criterion did not use phase or speed estimation performance.

Available treadmill settings constrained online calibration to 0.75, 1.00, and 1.25 m/s. To approximate the all-speed baseline under this constraint, the reference-subject mean-squared baseline error was minimized by
\begin{equation}
b_{new}=0.133x_{0.75}+0.252x_{1.00}+0.615x_{1.25}.
\end{equation}
These weights were determined from the 20-subject reference analysis rather than tuned on the online participant. Thus, the offline baseline used an equal average selected for reconstruction accuracy, whereas the online baseline used reference-derived weights for the available treadmill speeds. The two procedures addressed the same objective---approximating an all-speed user baseline---under different calibration-speed constraints. Table \ref{tab:calibration_speeds} briefly shows the calibration settings for offline and online evaluations.

\begin{table}[htb]
\renewcommand{\arraystretch}{1.25}
\caption{\label{tab:calibration_speeds}Calibration settings for offline and online analyses.}
\begin{center}
\begin{tabular*}{\columnwidth}{@{\extracolsep{\fill}}l c >{\raggedright\arraybackslash}p{0.48\columnwidth}@{}}
\hline
Setting & Speeds (m/s) & Baseline construction \\
\hline
Offline & 0.90, 1.15, 1.35 & Equal average; speeds selected by baseline reconstruction RMSE \\
Online & 0.75, 1.00, 1.25 & Weighted average for available treadmill speeds \\
\hline
\end{tabular*}
\end{center}
\end{table}

\subsection{Personalized Kernel Generation and Comparison Methods}

After constructing the calibration baseline, the personalized kernel library was regenerated by adding the reference PC1 speed deformation to the new user's baseline. Three methods were compared offline and online: personalized kernels with (A) and without (B) a PC1 anchor offset, and the population kernel (C).

The anchor offset is not the intercept of the quadratic regression. At anchor speed $v_a$, the user's calibration trajectory was first centered by $b_{new}$ and projected onto the reference PC1 direction to obtain $a_{new}(v_a)$. The offset is the difference between this projected score and the regression prediction $\hat{a}(v_a)$:
\begin{equation}
\delta a=a_{new}(v_a)-\hat{a}(v_a).
\end{equation}
\begin{equation}
\begin{aligned}
\hat{K}^{A}_{new,v}&=b_{new}+\left(\hat{a}(v)+\delta a\right)p_1,\\
\hat{K}^{B}_{new,v}&=b_{new}+\hat{a}(v)p_1.
\end{aligned}
\end{equation}
Method A adds this subject-specific correction to the predicted score at every candidate speed, preserving the regression-predicted speed dependence while shifting its PC1-score level. Method B sets $\delta a=0$ and therefore personalizes only the baseline trajectory. Here, $v_a$ was 1.15 m/s offline and 1.00 m/s online; Method C used the population-average reference kernel without user calibration.

\section{Experimental Setup}

\subsection{Offline Validation}

Methods A--C were evaluated on the 20-subject, 0.50--1.85 m/s dataset using LOSO cross-validation and the fixed calibration speeds above. In each fold, the other 19 subjects defined the population kernel and PCA basis. The quadratic form was held fixed, but its coefficients were re-estimated using the same 19 subjects. The held-out subject provided only the fixed-speed calibration trajectories and the trajectories used for phase and speed evaluation.

\subsection{Online Validation}

The online validation used steady-state, constant-speed walking trials collected from one male participant (29 years, 168 cm, 65 kg). Two MPU9250 IMUs were attached to the left and right thighs, two ICM20948 IMUs were attached to the left and right shanks, and two force-sensitive resistors (FSRs) were attached beneath the bilateral heels to provide gait-event references. Each IMU's sensing axes were aligned with the sagittal plane, and quiet standing segment angles were zeroed before each measurement. A Jetson Nano sampled raw accelerometer and gyroscope data at 100 Hz, estimated sagittal plane segment angles using an adaptive complementary filter whose fusion weight responded to acceleration and angular velocity magnitudes to reduce dynamic acceleration effects, and executed online matching.

Calibration and validation data were collected in separate sessions so that the trajectories used to construct the personalized baseline were not reused as validation trials. Calibration comprised 30-s trials at 0.75, 1.00, and 1.25 m/s; validation comprised separate 30-s trials at 0.50, 0.75, 1.00, 1.25, 1.50, and 1.75 m/s. All trials were constant-speed and excluded dynamic transitions.

\begin{figure}[thb]
\begin{center}
\includegraphics[width=0.5\columnwidth]{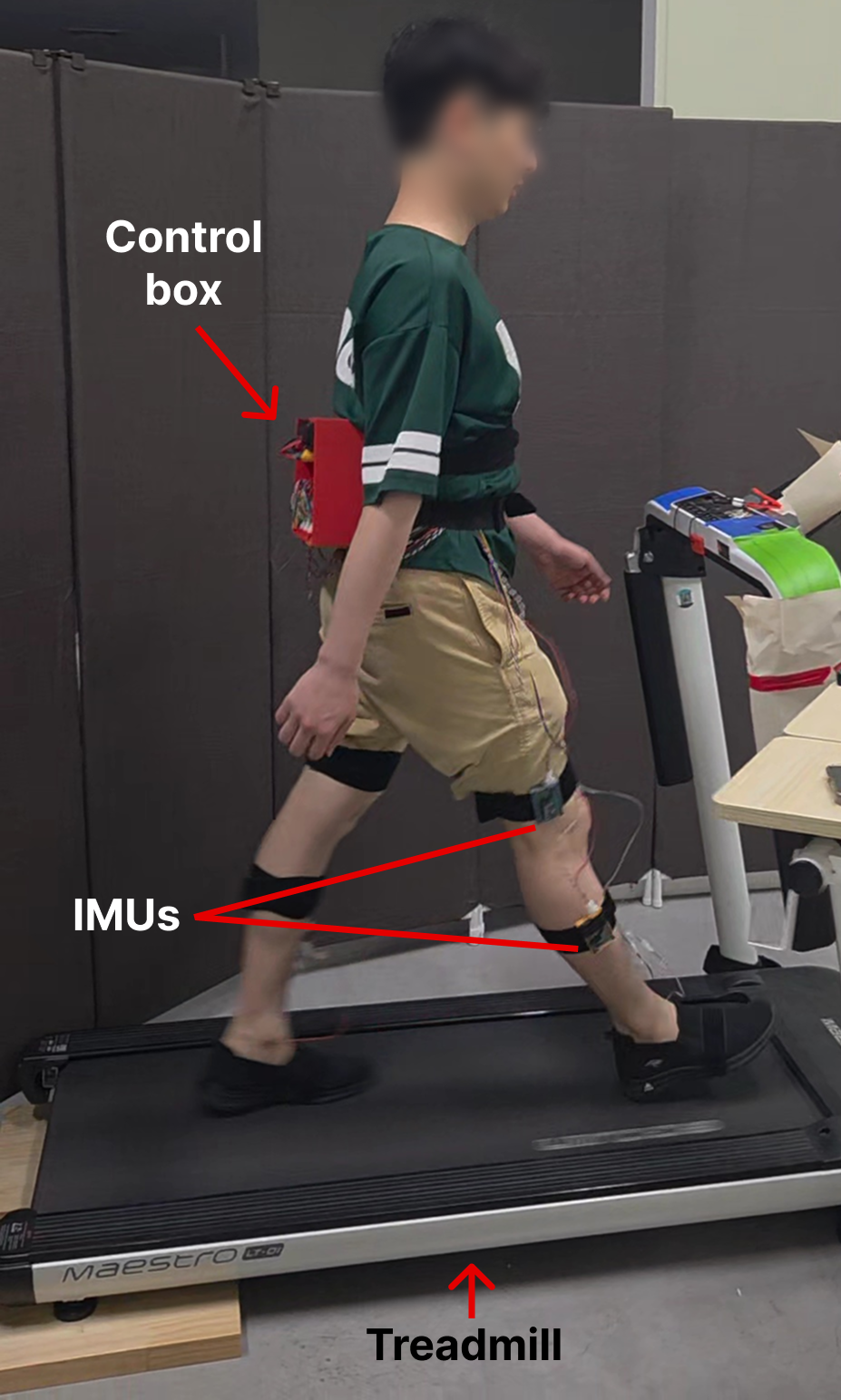}
\caption{\label{fig:online_setup}Online setup with four thigh/shank IMUs, heel FSRs, and a Jetson Nano for 100-Hz acquisition and matching.}
\end{center}
\end{figure}

\subsection{Evaluation Metrics and Statistical Analysis}

Phase estimation error was evaluated using circular phase error. The signed phase error was wrapped to the interval $[-0.5,0.5)$, and phase RMSE was computed from the wrapped error. Speed RMSE used treadmill speed as ground truth.

For offline validation, Method A, Method B, and Method C were compared using a subject-level repeated-measures design. Method effects were evaluated using a Friedman test across subjects, and pairwise Wilcoxon signed-rank tests were corrected using the Holm method.

For the single-participant online validation, stride-level RMSE values were first averaged within each treadmill speed and method. The resulting speed-level mean RMSEs were compared among Methods A, B, and C using a Friedman test with treadmill speed as the blocking factor; significant method effects were followed by Holm-corrected pairwise Wilcoxon signed-rank tests. The same comparison was repeated after excluding the 0.50 m/s lower-boundary condition as a sensitivity analysis.

\section{Results}

\subsection{Offline Validation Results}

Table~\ref{tab:offline_results} shows that Method A reduced mean phase RMSE from 0.0178 to 0.0161 compared with the non-personalized population kernel, corresponding to a 9.55\% reduction. The Friedman test showed a significant method effect for offline phase RMSE ($\chi^2=15.10$, $p=0.0005$). In the Holm-corrected Wilcoxon comparisons, Method A and Method B both reduced phase RMSE compared with Method C ($p=0.0005$ and $p=0.0017$, respectively), whereas the difference between Method A and Method B was not significant ($p=0.7285$). The speed RMSE was similar across the three methods: Method A achieved 0.1104 m/s, Method B achieved 0.1103 m/s, and the non-personalized population kernel achieved 0.1118 m/s.

\begin{table}[htb]
\setlength{\extrarowheight}{0.75ex}
\caption{\label{tab:offline_results}Offline validation results, reported as subject-level mean $\pm$ standard deviation.}
\begin{center}
\begin{tabu}to\columnwidth{X[c]X[c]X[c]}\hline
Method & Phase RMSE & Speed RMSE\\
\hline
A & $0.0161\pm0.0040$ & $0.1104\pm0.0625$\\
B & $0.0161\pm0.0039$ & $0.1103\pm0.0637$\\
C & $0.0178\pm0.0045$ & $0.1118\pm0.0581$\\
\hline
\end{tabu}
\end{center}
\end{table}

\subsection{Online Validation Results}

\begin{figure}[thb]
\begin{center}
\includegraphics[width=\columnwidth]{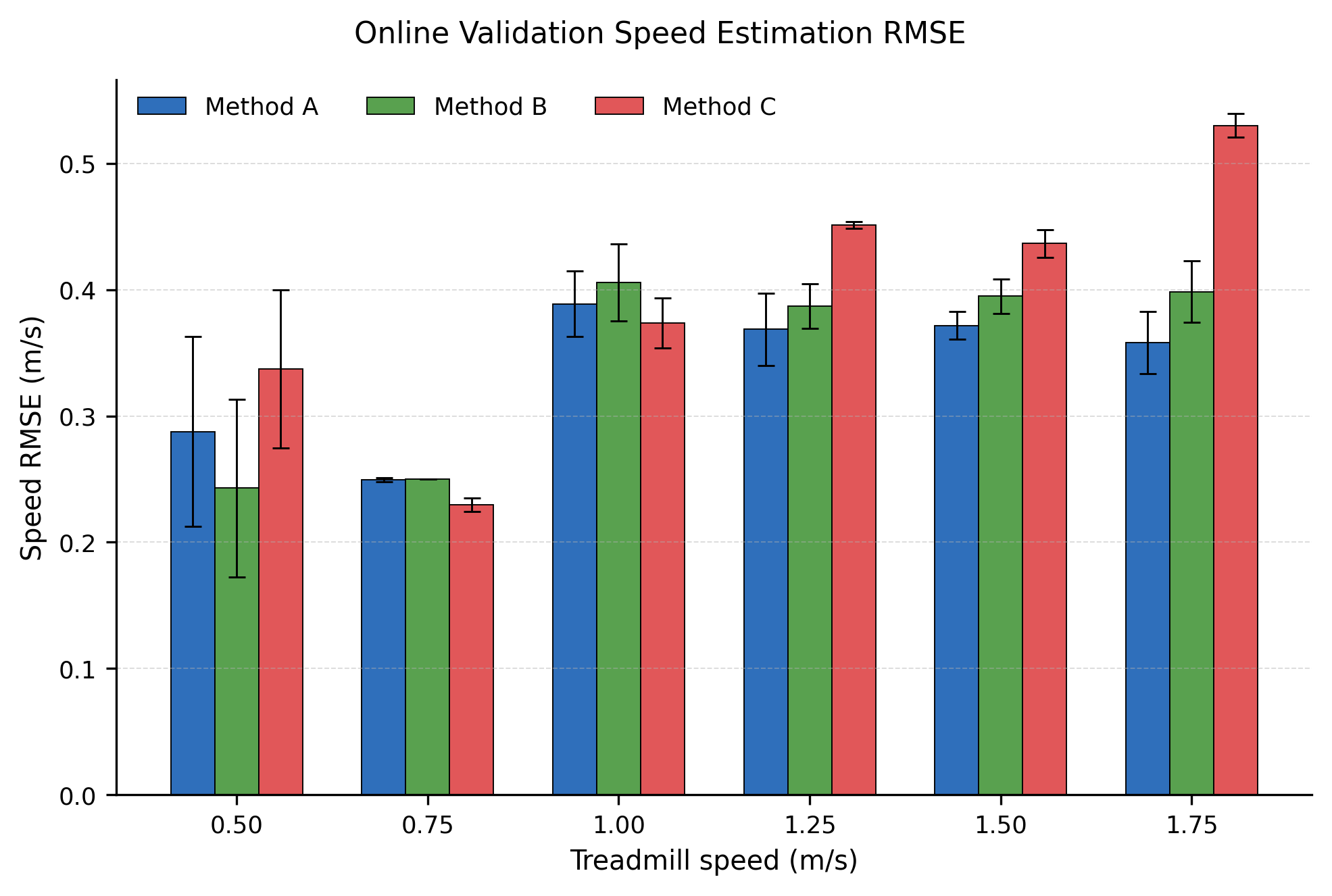}
\caption{\label{fig:online_speed_rmse}Stride-level online speed estimation RMSE (within-subject mean $\pm$ SD).}
\end{center}
\end{figure}

\begin{figure}[thb]
\begin{center}
\includegraphics[width=\columnwidth]{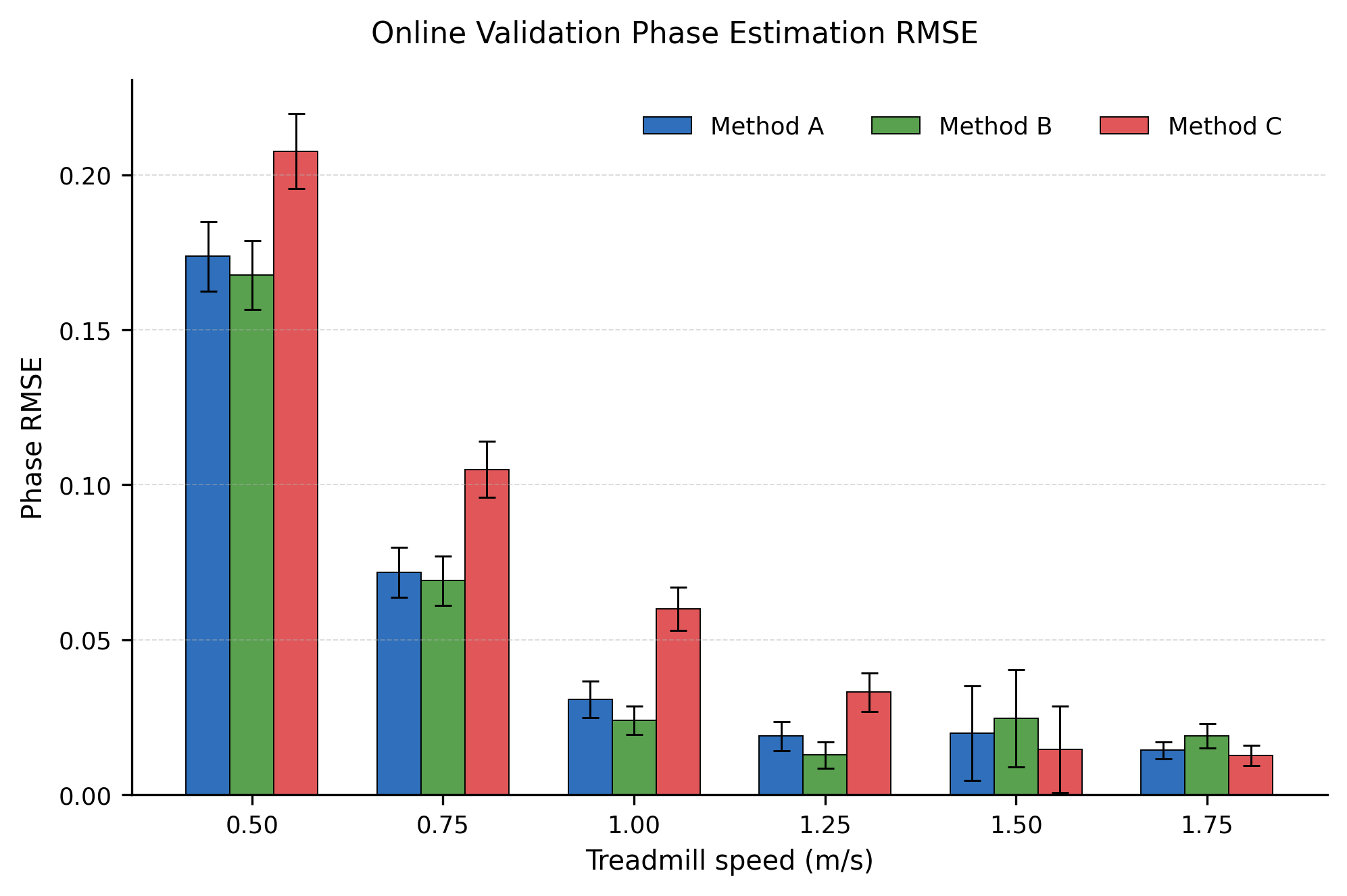}
\caption{\label{fig:online_phase_rmse}Stride-level online phase estimation RMSE (within-subject mean $\pm$ SD).}
\end{center}
\end{figure}

The same incoming IMU angle trajectories were evaluated with Method A, Method B, and Method C. Across the six online validation speeds, Method A showed the lowest mean speed RMSE (0.3349 m/s), followed by Method B (0.3451 m/s) and Method C (0.3913 m/s). The corresponding mean phase RMSE values were 0.1316 for Method A, 0.1279 for Method B, and 0.1503 for Method C. The speed RMSE difference was not significant in the single-subject blocked comparison (Friedman $\chi^2=2.33$, $p=0.3114$). For phase RMSE, the Friedman test indicated a method effect ($\chi^2=10.33$, $p=0.0057$). Pairwise Wilcoxon tests suggested lower phase RMSE for Methods A and B than for Method C before correction ($p=0.0313$ for both), but these comparisons did not remain significant after Holm correction (all $p_{\mathrm{Holm}}=0.0938$). Excluding the 0.50 m/s lower-boundary condition gave the same interpretation: speed RMSE was not significant ($p=0.2466$), and phase RMSE retained a Friedman method effect ($p=0.0150$) without significant Holm-corrected pairwise differences. Fig.~\ref{fig:online_speed_rmse} shows the speed estimation RMSE, and Fig.~\ref{fig:online_phase_rmse} shows the phase estimation RMSE. Each bar represents the mean across strides at one treadmill speed for the single online participant, and each error bar represents the corresponding within-subject stride-level standard deviation. For the high-speed trials in which FSR over-detection was verified, corrected right-heel-strike event logs were used only for phase ground-truth reconstruction; the IMU inputs and online estimator outputs were unchanged.

\subsection{Embedded Runtime}

The matching computation was fast enough for real-time embedded use. Across the six 30-s online walking trials, the average matching time per update was 48.49 $\mu$s for Method A, 49.03 $\mu$s for Method B, and 48.89 $\mu$s for Method C. Running all three methods simultaneously required 146.41 $\mu$s per update, corresponding to approximately 1.73--1.75 $\mu$s per kernel for the 28-speed library.

\begin{table}[htb]
\renewcommand{\arraystretch}{1.25}
\caption{\label{tab:latency}Online matching latency.}
\begin{center}
\begin{tabular*}{\columnwidth}{@{\extracolsep{\fill}}lcc@{}}
\hline
Method & Total time ($\mu$s) & Per kernel ($\mu$s) \\
\hline
A & $48.49 \pm 0.77$ & 1.73 \\
B & $49.03 \pm 0.84$ & 1.75 \\
C & $48.89 \pm 0.95$ & 1.75 \\
A+B+C & $146.41 \pm 2.51$ & -- \\
\hline
\end{tabular*}
\end{center}
\end{table}

\section{Discussion}

\subsection{Phase and Speed Estimation Performance}

The offline results demonstrate that sparse calibration can improve phase estimation within the motion-capture-derived reference domain. Across 20 subjects, personalized kernels reduced phase RMSE relative to the population kernel, whereas speed RMSE remained similar. The absence of a difference between Methods A and B indicates that the personalized baseline accounted for most of the observed offline phase benefit; the additional anchor correction was not consistently advantageous.

Online, the personalized kernels showed lower descriptive mean phase and speed RMSEs across six speeds. The overall phase method effect and uncorrected pairs suggested a possible benefit, but corrected pairwise phase tests and the speed test were not significant. These single-participant results therefore provide within-subject proof of concept, not population-level inference. Phase errors were lowest near 1.00--1.25 m/s and larger at 0.50 m/s and at high speeds. The 0.50 m/s result likely reflects a lower-boundary effect: when the true speed is already the lowest kernel speed, no slower kernel can absorb residual mismatch. Because the same selected kernel determines both outputs, this mismatch can affect speed and phase together.

\subsection{Reference-to-Online Dataset Mismatch}

The residual speed underestimation was not explained well by simple IMU amplitude attenuation, because the measured IMU range of motion was not consistently smaller than the motion-capture-derived kernels. The remaining error is more plausibly explained by a broader reference-to-online dataset mismatch. The PC basis and PC1-speed regression were learned from the open-source reference dataset, whereas the online measurements were collected separately from a new participant, with different hardware, attachment conditions, walking sessions, and preprocessing. The fact that the reference trajectories were motion-capture-derived and the online trajectories were wearable-IMU-derived is one important part of this mismatch, but it is not the only possible source.

\begin{figure}[thb]
\begin{center}
\includegraphics[width=\columnwidth]{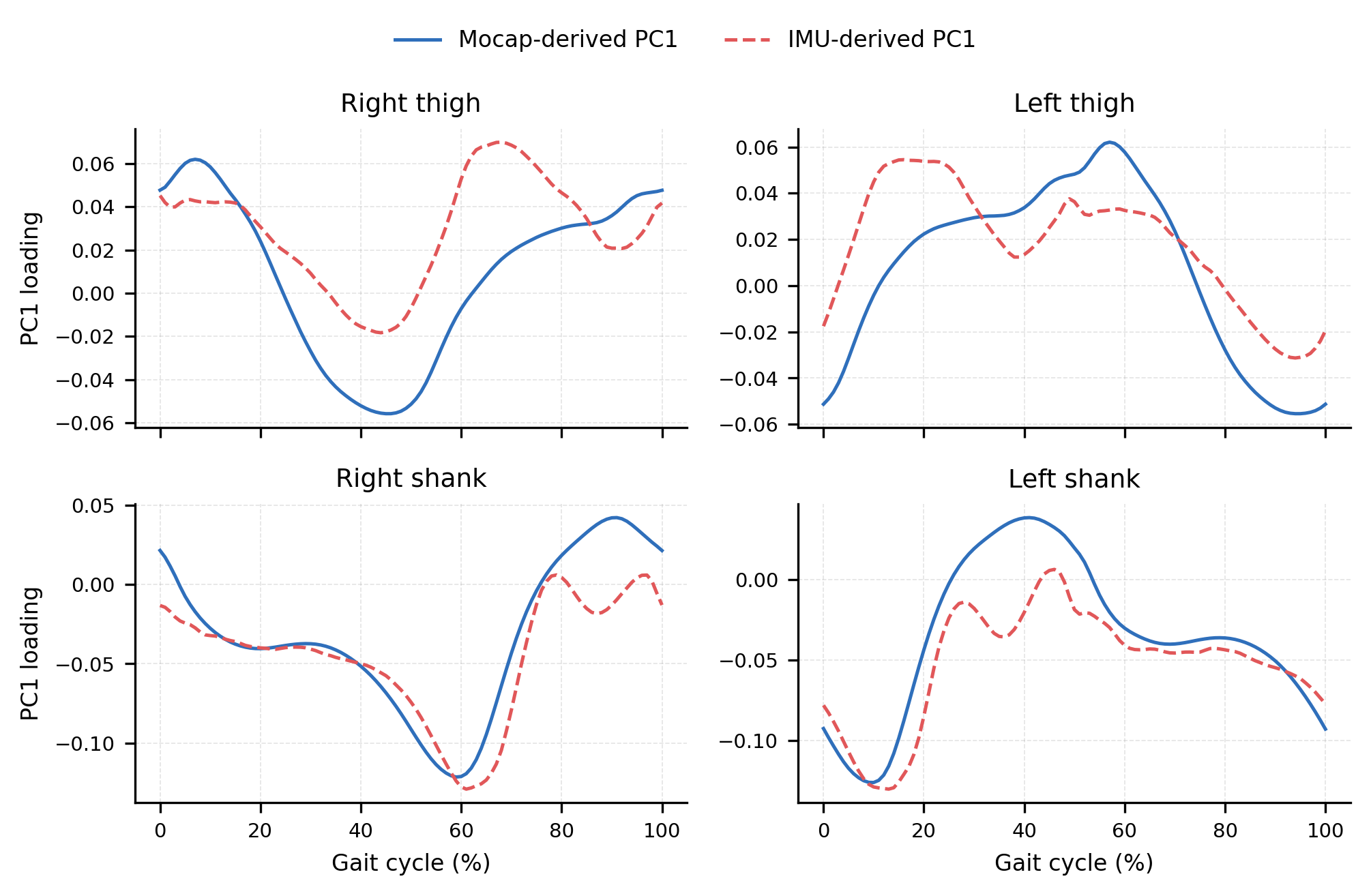}
\caption{\label{fig:imu_vs_mocap_pc}Comparison between motion-capture-derived PC1 and IMU-derived PC1 waveforms for the four kinematic channels. Each subplot shows the PC1 loading over the gait cycle for one segment.}
\end{center}
\end{figure}

When PC was recomputed from the collected IMU angle data using the same baseline-centered procedure, the sign of the resulting IMU-domain PC1 was aligned with that of the motion-capture-derived PC1 before comparison. After alignment, the global four-channel cosine similarity was 0.841 (Fig.~\ref{fig:imu_vs_mocap_pc}), indicating that the speed-dependent deformations were similar but not identical across the two datasets. In addition to the sensing modality, subject-specific IMU placement and alignment can change the measured thigh and shank angle waveforms. Therefore, the online performance degradation should be interpreted as the combined effect of cross-dataset shift, sensor-domain differences, and wearable-sensor attachment variability.

\subsection{Limitations}

Several limitations should be noted. First, personalized kernels combined an IMU-derived calibration baseline with a PC model learned from a separate reference dataset rather than entirely from IMU-domain data. Consequently, personalization corrected the user's baseline shape but retained a speed-deformation direction learned from motion-capture-derived reference trajectories. Second, the online study included only one participant and cannot establish generalization across users, sensor placements, or environments. The speed-blocked analysis quantifies within-participant differences among methods but does not replace a participant-level population analysis.

Third, no explicit IMU placement calibration was evaluated. Because attachment position and orientation may vary across users and sessions, such calibration may reduce waveform mismatch before matching. Fourth, calibration speeds and regression order were selected on the same 20-subject dataset used for LOSO evaluation. Although selection did not optimize phase or speed RMSE directly, it was not nested and may introduce dataset-specific optimization. Offline and online RMSEs are also not directly comparable because calibration speeds, baseline construction, sensing domains, and protocols differed. Finally, the online evaluation and kernel library were limited to steady-state walking; performance during speed transitions, gait initiation, and termination remains unknown.

Future work should address two priorities. First, larger participant cohorts and sufficient wearable data are needed to repeat baseline-centered PCA, derive IMU-domain PCs and PC1-speed regressions, and evaluate explicit placement calibration. Second, calibration-speed combinations and regression order should be selected using nested subject-level validation or an independent development dataset. Subsequent evaluation should then include dynamic speed changes and gait initiation and termination to determine whether a steady-state kernel library remains adequate under transient conditions.

\section{Conclusion}

This study generated personalized 28-speed, four-channel kernels from three calibration speeds and a reference-derived PC1 deformation model. Personalization significantly reduced offline phase RMSE while leaving speed RMSE similar. The single-participant pilot produced numerically lower online phase and speed errors and demonstrated real-time embedded feasibility, but speed differences and corrected pairwise phase differences were not significant. By changing the kernel-generation stage rather than the online search procedure, the method preserves the simplicity and low latency of sliding-kernel estimation. The results support sparse-calibration feasibility while motivating larger-cohort validation with IMU-domain modeling, placement calibration, and transient walking conditions.

\section*{Acknowledgment}

This work was supported by the National Research Foundation of Korea (NRF) grant funded by the Ministry of Science and ICT (MSIT), Republic of Korea (No. RS-2026-25492395).

\bibliographystyle{IEEEtran}
\bibliography{references}

\end{document}